\documentclass[10pt]{article}

\usepackage[margin=0.9in]{geometry}
\usepackage[T1]{fontenc}
\usepackage{amsmath,amssymb}
\usepackage{booktabs}
\usepackage{graphicx}
\usepackage{microtype}
\usepackage[numbers,sort&compress]{natbib}
\usepackage{multirow}
\usepackage{float}
\usepackage{placeins}
\usepackage[hidelinks]{hyperref}

\newcommand{\Dmax}{D_{\max}}
\newcommand{\cN}{\mathcal N}

\title{Population Scaling or Data Dilution?\\
Dynamics of Local Topology Evolution in Decentralized Learning}
\author{Yin-Kuan Liang$^{1}$ \quad Yan Gao$^{2}$ \quad Yang Long$^{1}$\\
$^{1}$Durham University $^{2}$University of Cambridge}
\date{}

\begin{document}
\maketitle

\begin{abstract}
Scaling decentralized learning changes not only the number of clients \(N\), but also the dynamics of information propagation and consensus. We argue that the effect of increasing \(N\) cannot be understood in isolation, because data allocation, topology-dependent mixing, and communication capacity may change simultaneously. We study these coupled effects on CIFAR-10 with \(N\in\{10,50,100,200\}\), comparing a degree-two Ring, a Static Random graph, and Local-First Heuristic Evolution (LFHE), a locally adaptive topology process based on friend-of-friend discovery. The Ring provides an analytically transparent failure mode: its Metropolis spectral gap decays as \(\Theta(N^{-2})\), implying progressively slower contraction of model disagreement as the population grows. Experiments show that holding the nominal local dataset size fixed substantially reduces the apparent population penalty observed when a fixed total dataset is divided among more clients. Because our protocol uses one local epoch per round, this comparison also changes local SGD workload and cannot isolate data availability as the sole cause. The remaining degradation depends strongly on communication structure: Ring enters a high-disagreement regime, whereas Static Random and LFHE remain close to consensus. Increasing LFHE's degree threshold further improves accuracy and consensus, but at a substantially higher model-transmission cost. These results show that decentralized scaling is governed by coupled learning and communication dynamics, rather than by the number of clients alone.
\end{abstract}

\section{Introduction}

Decentralized learning replaces server aggregation with peer-to-peer model exchange, building on distributed optimization over communication graphs~\citep{tsitsiklis1986distributed,nedic2009subgradient,lian2017dpsgd,koloskova2020compression}. 
Without a central coordinator, optimization becomes directly coupled to the communication topology: the graph structure and mixing weights determine how rapidly information propagates and local models approach consensus~\citep{boyd2006gossip,olfati2007consensus,nedic2018topology}.
When the topology itself adapts during training, this dependence becomes bidirectional. Model states influence which edges are created or removed, the resulting graph determines subsequent mixing, and mixing in turn changes the model states that guide future topology updates.

Understanding how such systems scale with the number of clients \(N\) is difficult because several resources may change simultaneously. Under a fixed total dataset, increasing \(N\) enlarges the communication network while reducing the amount of data assigned to each client.
Adaptive sparse graphs introduce a second scaling dimension: if their admissible degree grows with \(N\), larger populations also receive greater communication and mixing capacity. Consequently, an observed scaling curve may reflect data allocation, topology-dependent mixing, communication capacity, or a combination of all three. Distinguishing these effects is essential because the convergence of decentralized optimization depends directly on the nontrivial spectrum of the communication operator~\citep{boyd2004fastest,xiao2004fast,boyd2006gossip,olfati2007consensus}.\looseness-1

Recent work has explored adaptive peer selection and communication topology under heterogeneous data, including performance-based neighbor selection~\citep{onoszko2021pens} and diversity-driven rewiring~\citep{cox2026morph}. Local-First Heuristic Evolution (LFHE) was introduced as a bounded, representation-driven FoF topology-search mechanism~\citep{liang2026lfhe}. Unlike methods using expanding peer-discovery state or online global spectral objectives, LFHE samples local two-hop candidates and evaluates initiator-local add-or-swap operations. The companion LFHE preprint provides the mechanism and multi-task evaluation; the present paper specifically examines its population-, data-, and degree-scaling dynamics. Our focus is complementary: we study how population size, data allocation, and communication capacity jointly shape the dynamics of decentralized learning. We compare three communication structures: a degree-two Ring, which provides an analytically transparent mixing bottleneck; a Static Random graph, which serves as a fixed sparse reference; and LFHE, a locally adaptive topology process based on two-hop friend-of-friend (FoF) discovery. For LFHE, we separately study a fixed degree threshold and a population-scaled threshold, allowing us to distinguish the effects of population growth from those of increasing communication capacity.

Our contributions are threefold. First, we show that the apparent population penalty depends strongly on the data-allocation and local-optimization protocol (which are coupled by our one-epoch schedule). For LFHE, the accuracy loss from \(N=50\) to \(N=200\) decreases from \(14.14\) to \(3.68\) percentage points when moving from a fixed-total regime to one with a fixed nominal local dataset size. Second, we characterize a locality--communication trade-off: uniformly bounded realized degree limits neighbor exposure, but causes the fraction of the population visible through two-hop discovery to shrink with \(N\); increasing the degree threshold expands both local search and mixing capacity, at a rapidly growing communication cost. Third, topology, model-disagreement, and communication-normalized trajectories reveal how these resource choices produce distinct graph--model regimes. Together, these results shift the scaling question from \emph{how accuracy changes with \(N\)} to \emph{which data and communication resources must scale with \(N\) to preserve effective decentralized learning}.

\section{Local Topology Evolution Under Scaling}
\label{sec:method}

Let \(G_t=(V,E_t)\) denote the communication graph and \(W_t=(w_{1,t},\ldots,w_{N,t})\) the client models. Each communication round performs local optimization followed by graph-dependent mixing,
\begin{equation}
\widetilde W_t=\mathcal L(W_t;\xi_t),\qquad
W_{t+1}=P(G_t)\widetilde W_t .
\label{eq:model}
\end{equation}
Every \(H=5\) rounds, LFHE additionally applies \(G_{t+1}=\mathcal R_t(G_t,W_{t+1})\); otherwise the graph is unchanged. The learning system is therefore coupled: \(G_t\) determines model mixing, while \(W_t\) provides the local signals used to evolve the topology. We use the local-search framework of~\citet{liang2026lfhe}, but evaluate it here under controlled scaling regimes.

\begin{figure}[t]
\centering
\includegraphics[width=0.96\linewidth]{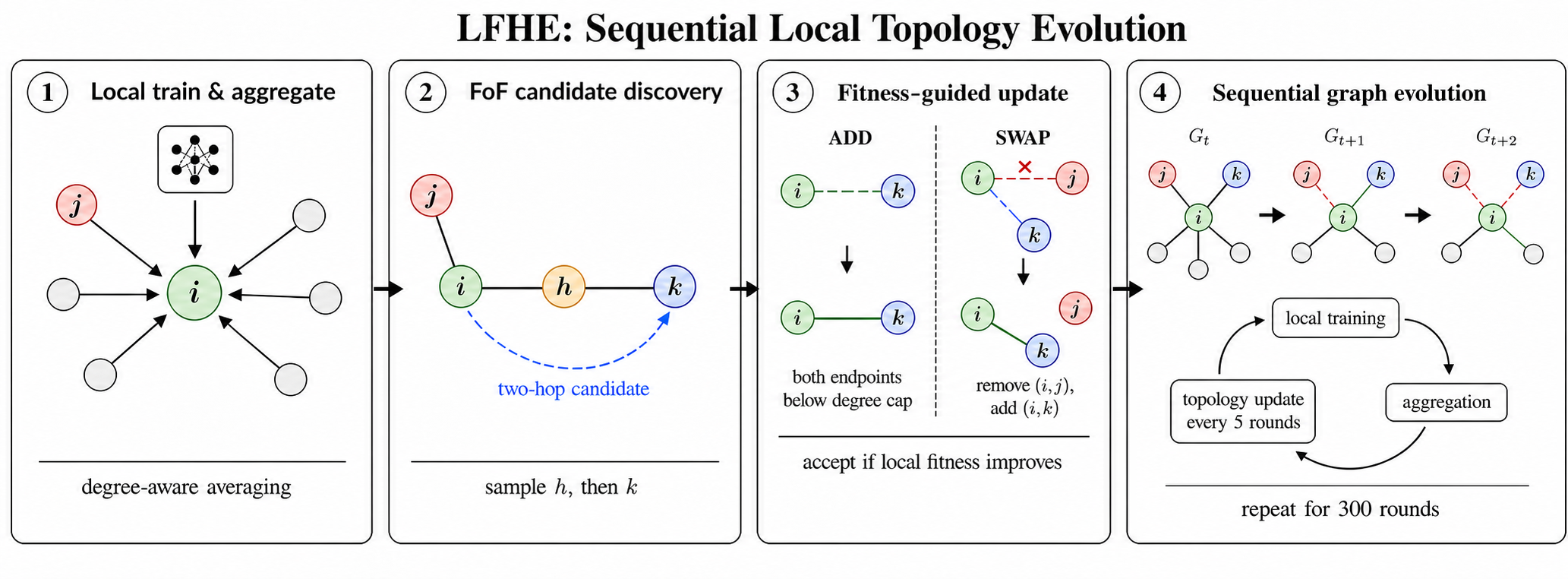}
\vspace{-5pt}
\caption{\small LFHE mechanism (following~\citet{liang2026lfhe}). Client \(i\) discovers a two-hop FoF candidate \(k\), evaluates a locally improving addition or edge replacement, and sequentially commits accepted updates.}
\label{fig:overview}
\vspace{-6pt}
\end{figure}

\paragraph{Local topology evolution.}
Client \(i\) evaluates
\begin{equation}
f_i=(1-\beta_t)C_i+\beta_tD_i-0.1\,d_i/\Dmax,
\qquad
\beta_t=e^{-0.01t},
\label{eq:fitness}
\end{equation}
where \(C_i=\sum_{j\in\cN_i}\|z_i-z_j\|_2^2\) scores incident-edge structural disagreement and \(D_i=1-\cos(z_i,|\cN_i|^{-1}\sum_{j\in\cN_i}z_j)\) measures mismatch with the mean neighbor representation. In these experiments, \(z_i\in\mathbb R^{10}\) is the class-wise mean of the last classifier's \(10\times256\) weights (Appendix~\ref{app:method}). Client \(i\) samples \(h\in\cN_i\), then \(k\in\cN_h\setminus(\cN_i\cup\{i\})\). It accepts an add or a feasible one-edge swap only if its own score strictly improves. Adds require both endpoints below \(\Dmax\); swaps use a common-neighbor redundancy check and the best improving eligible edge replacement. Within an epoch, one FoF proposal is considered per initiator in a fixed sequential order, so later initiators see earlier commits. \(\Dmax\) regulates online edge acquisition, not a strict maximum on initialized node degrees.

\paragraph{Local visibility.}
If \(\bar d_t=\max_i d_i(G_t)\), the two-hop candidate set satisfies
\begin{equation}
|\mathcal C_i^{(2)}(t)|
\le d_i(t)(\bar d_t-1),\qquad
\frac{|\mathcal C_i^{(2)}(t)|}{N-1}
\le
\frac{\bar d_t(\bar d_t-1)}{N-1}.
\label{eq:coverage}
\end{equation}
If the \emph{realized} maximum degree \(\bar d_t\) stays uniformly bounded as \(N\) grows, relative FoF visibility therefore shrinks as \(O(N^{-1})\). For illustration only, in a graph with an actual maximum degree of four, any client has at most 12 distinct two-hop candidates (\(12/49=24.5\%\) at \(N=50\), \(12/99=12.1\%\) at \(N=100\)). These values are \emph{not} guaranteed by the online threshold \(\Dmax=4\), because the initial connected random graph can contain nodes of higher degree. The bound formalizes a conditional locality--coverage tension, not a hard candidate cap for every experimental snapshot.

\paragraph{Mixing and disagreement.}
For symmetric Metropolis--Hastings mixing \(P_t=P(G_t)\), let \(J=\frac{1}{N}\mathbf{1}\mathbf{1}^{\top}\) denote global averaging and define

$$
\rho_t=\|P_t-J\|_2.
$$

Then one communication step satisfies
\begin{equation}
\lVert(I-J)P_tX\rVert_F \le \rho_t\lVert(I-J)X\rVert_F,
\label{eq:contraction}
\end{equation}
so \(\rho_t\) directly controls how quickly model disagreement contracts: smaller \(\rho_t\) implies faster mixing~\citep{boyd2004fastest,xiao2004fast,boyd2006gossip,olfati2007consensus}. Under non-IID data, local optimization continually reintroduces disagreement, making the learning dynamics a balance between local drift and network mixing. Our reported algebraic connectivity \(\lambda_2(L_t)\) uses the \emph{unweighted} graph Laplacian and is an offline structural diagnostic; it is not the Metropolis contraction factor \(\rho_t\) and is not optimized online by LFHE.

This dependence becomes explicit for a degree-two Ring, whose mixing eigenvalues are
\begin{equation}
\mu_k=\frac{1+2\cos(2\pi k/N)}{3},\qquad 1-\mu_1=\frac{2}{3}\left(1-\cos\frac{2\pi}{N}\right)=\Theta(N^{-2}).
\label{eq:ringgap}
\end{equation} 
Intuitively, information must travel around an increasingly long cycle, so local updates are averaged across the population progressively more slowly.
Hence, as \(N\) grows, the Ring spectral gap vanishes quadratically and disagreement contracts increasingly slowly. The Ring therefore provides a concrete population-dependent mixing failure mode. Static Random offers a contrasting sparse regime with approximately constant local degree but without the deterministic cycle bottleneck.

\section{Controlled Population and Degree Scaling}
\label{sec:experiment}

We use CIFAR-10~\citep{krizhevsky2009learning} with class-wise Dirichlet heterogeneity \(\alpha=0.3\), \(N\in\{10,50,100,200\}\), and seeds \(\{42,43,44\}\). Training runs for 300 communication rounds with one local epoch per round, SGD with learning rate \(0.05\), and batch size \(32\). Evaluation and LFHE topology updates occur every five rounds. We report means and sample SD; further details appear in Appendix~\ref{app:details}.

We compare \textbf{Ring}, a fixed cycle; \textbf{Static Random}, a fixed connected sparse graph; and \textbf{LFHE}. Static Random and LFHE share seed-matched connected Erd\H{o}s--R\'enyi initializations with expected degree four; all methods use degree-aware Metropolis--Hastings mixing.

\paragraph{Data and degree regimes.}
Under \emph{fixed-total}, the 50,000 CIFAR-10 training examples are partitioned across the population, so the mean per-client allocation decreases with \(N\). Under \emph{fixed-per-client}, each client instead receives 500 sampled local indices, keeping the \emph{nominal} number of entries fixed. Indices are reused if a client's original pool contains fewer than 500 distinct examples. Since we use one local epoch, the two regimes also change the number of local SGD steps; the comparison is a joint data/training-protocol contrast, not a causal isolation of distinct data quantity (Appendix~\ref{app:details}).

LFHE uses either a fixed online threshold \(\Dmax=4\) or the population-dependent schedule
\[
\Dmax=\max(2,\lfloor2\ln N\rfloor)=\{4,7,9,10\}.
\]
The fixed-threshold setting tests population growth under approximately constant local communication capacity; the scaled-threshold setting measures the benefit and cost of increasing that capacity with population.

Model disagreement and cumulative directed model transmissions are
\begingroup
\small
\begin{equation}
\mathcal D_t=
\frac{1}{N}\sum_{i=1}^N
\left\lVert
\theta_{i,t}-\frac{1}{N}\sum_j\theta_{j,t}
\right\rVert_2^2,
\qquad
M(T)=2\sum_{t<T}|E_t|.
\label{eq:metrics}
\end{equation}
\endgroup
Communication accounting includes directed model exchange; topology-control traffic, transmitted bytes, and wall-clock latency are not included in this metric. Consequently, matching these transmission counts does not guarantee equal end-to-end communication cost.

\section{What Drives Population Scaling?}
\label{sec:results}

\begin{figure}[t]
\centering
\includegraphics[width=0.9\linewidth]{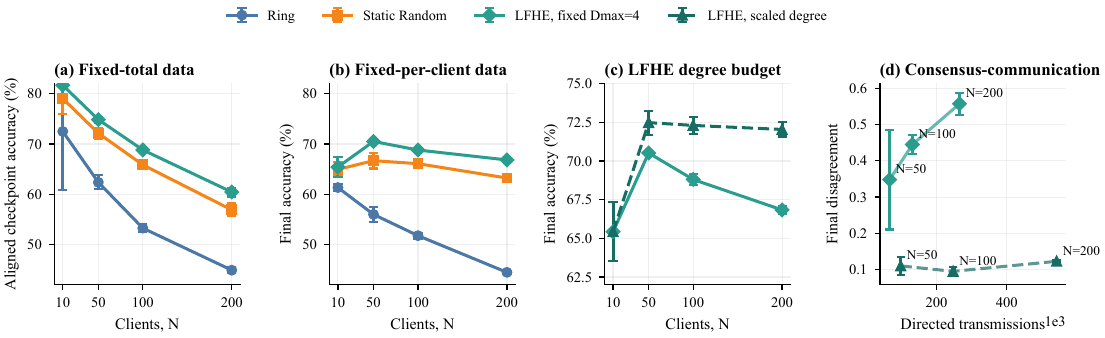}
\vspace{-5pt}
\caption{\small Controlled scaling on CIFAR-10. (a) Fixed-total population scaling; (b) fixed-per-client scaling with \(\Dmax=4\); (c) LFHE with fixed and population-scaled degree thresholds; (d) disagreement versus realized directed model transmissions. Panel (a) uses the aligned cross-method checkpoint, whereas panels (b)--(d) report final endpoints. Error bars denote sample SD across seeds 42--44.}
\label{fig:results}
\vspace{-6pt}
\end{figure}

\paragraph{Data/training protocol changes the scaling regime.}
With fixed-total data, fixed-\(\Dmax\) LFHE drops from \(74.65\%\) to
\(68.64\%\) over \(N=50\rightarrow100\) and from \(68.64\%\) to
\(60.51\%\) over \(N=100\rightarrow200\), losses of \(6.01\) and
\(8.13\) points. Under fixed-per-client training, the corresponding
losses are \(1.71\) points (\(70.52\%\rightarrow68.81\%\)) and \(1.97\)
points (\(68.81\%\rightarrow66.84\%\)). Under the fixed-nominal-per-client protocol, the apparent penalties are smaller by \(4.30\) and \(6.16\) points, respectively. The effect combines differences in allocated examples, repeated sampling, and local SGD steps; our runs do not separate their individual contributions.

\paragraph{Residual scaling depends on communication structure.}
Under fixed-per-client training, Ring falls from \(61.34\%\) to \(44.46\%\) over \(N=10\rightarrow200\), while Static Random remains comparatively stable (\(64.95\%\rightarrow63.24\%\)). Fixed-\(\Dmax\) LFHE reaches \(66.84\%\) at \(N=200\), \(3.60\) points above Static Random with approximately \(9.7\%\) more model transmissions. Thus endpoint differences alone do not isolate the contribution of adaptive edge selection from additional connectivity. Disagreement trajectories distinguish the same regimes: Ring moves toward high disagreement, whereas Static Random and LFHE remain near consensus (Figure~\ref{fig:support}a); topology trajectories are reported in Appendix~\ref{app:dynamics}.

\paragraph{Communication capacity sets a second scaling frontier.}
Increasing LFHE's degree threshold raises accuracy from \(70.52\%\) to \(72.49\%\) at \(N=50\), from \(68.81\%\) to \(72.31\%\) at \(N=100\), and from \(66.84\%\) to \(72.04\%\) at \(N=200\), while model transmissions increase by \(47.1\%\), \(88.1\%\), and \(104.1\%\), respectively (Appendix~\ref{app:degree}). At a common \(115.8\)k-transmission budget, fixed-\(\Dmax\) LFHE reaches \(67.77\%\), compared with \(65.98\%\) for Static Random and \(63.58\%\) for scaled-threshold LFHE; the latter attains its higher endpoint only after approximately \(247\)k transmissions (Figure~\ref{fig:support}b). Additional degree therefore extends the attainable learning regime at increasing communication cost.

\begin{figure}[t]
\centering
\includegraphics[width=0.9\linewidth]{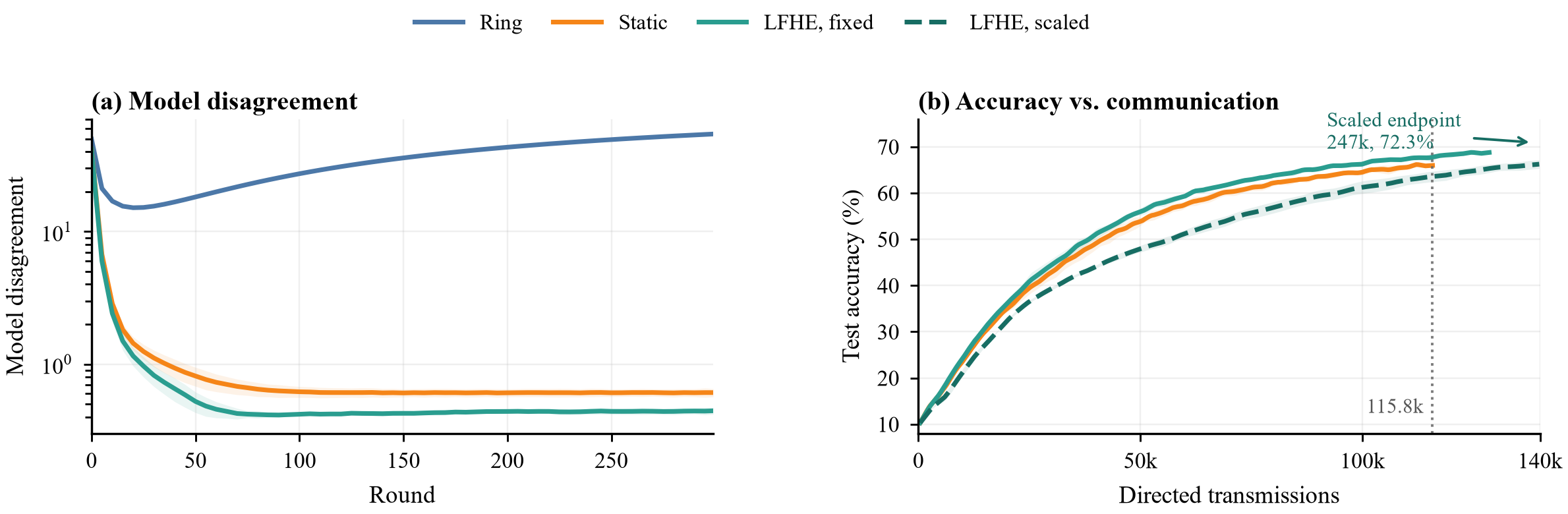}
\vspace{-5pt}
\caption{\small Dynamics at \(N=100\) under fixed-per-client training. (a) Ring enters a high-disagreement regime, while Static Random and fixed-\(\Dmax\) LFHE remain near consensus. (b) Communication-normalized accuracy: fixed-\(\Dmax\) LFHE leads within the shared budget, while scaled-threshold LFHE reaches a higher endpoint at greater cost.}
\label{fig:support}
\vspace{-6pt}
\end{figure}

\section{Discussion and Conclusion}

Two resource effects shape decentralized scaling. First, \textbf{data allocation together with local optimization}: changing the protocol reduces LFHE's observed losses from \(6.01\) to \(1.71\) points over \(N=50\rightarrow100\) and from \(8.13\) to \(1.97\) over \(N=100\rightarrow200\), revealing different behavior across structures. Ring deteriorates, Static Random remains comparatively stable, and LFHE maintains higher accuracy with low disagreement.

The second effect is \textbf{communication capacity}. Increasing LFHE's degree threshold improves mixing and raises the achievable endpoint, but the benefit comes with rapidly increasing model exchange. The fixed-threshold results show that LFHE retains a performance advantage over Static Random even when their realized communication is close, while the scaled-threshold results expose how additional degree can trade communication for accuracy.

Together, these findings suggest a broader view of decentralized scaling. Population size is not an isolated systems variable: its effect depends on how data, local training, topology, and communication resources scale with it. In the tested finite-population regime, the fixed-nominal-per-client protocol substantially reduces the observed scaling penalty, while topology and degree budget shape residual consensus and communication costs. Scaling studies should therefore report these resources jointly rather than summarize behavior through client count and endpoint accuracy alone.


\FloatBarrier

\begingroup
\footnotesize
\setlength{\bibsep}{0pt}
\bibliographystyle{plainnat}
\bibliography{references}
\endgroup


\clearpage
\appendix

\section{Implementation-Faithful LFHE}
\label{app:method}

For the controlled-scaling CIFAR-10 experiments, LFHE uses a 10-dimensional summary of the final classifier weight matrix as the client representation. Let \(W_i^t\in\mathbb{R}^{10\times256}\) denote client \(i\)'s final classifier weight matrix. The implemented representation is
\[
z_i^t
=
\frac{1}{256}W_i^t\mathbf{1}_{256},
\]
obtained by averaging each output class's 256 incoming weights. Its local terms are
\begin{align}
C_i(G,t)
&=
\sum_{j\in\cN_i(G)}
\lVert z_i^t-z_j^t\rVert_2^2,\\
D_i(G,t)
&=
1-\cos\!\left(
z_i^t,
\frac{1}{|\cN_i(G)|}
\sum_{j\in\cN_i(G)}z_j^t
\right),
\end{align}
with
\[
f_i(G,t)
=
(1-\beta_t)C_i(G,t)
+
\beta_tD_i(G,t)
-
0.1\frac{d_i(G)}{\Dmax},
\qquad
\beta_t=e^{-0.01t}.
\]

The structural term is representation-dependent and is not a direct estimator of algebraic connectivity. In particular, \(\frac12\sum_i C_i(G,t)=\mathrm{Tr}(Z_t^\top L(G)Z_t)\) is the unweighted representation Dirichlet energy of the current graph, as established for LFHE in~\citet{liang2026lfhe}. This is a state-dependent structural signal; the discrete Metropolis contraction factor is distinct from the unweighted Laplacian spectrum. At each topology epoch, initiators are processed once in a fixed sequential order. Each initiator samples
\[
h\sim\mathrm{Unif}(\cN_i),\qquad
k\in\cN_h\setminus(\cN_i\cup\{i\}),
\]
and evaluates that FoF candidate against its current graph state. A simple addition is available when the relevant online degree conditions are satisfied; otherwise, LFHE evaluates feasible one-edge replacements. Each initiator commits at most one accepted add-or-swap operation before processing moves to the next client, which then observes the already updated graph. The implementation therefore uses sequential rather than concurrent topology commits.

\paragraph{Serialized topology-update pseudocode.}
\begin{enumerate}
\setlength{\itemsep}{0pt}
\setlength{\parsep}{0pt}
\item Copy the current graph and freeze client representations for this update epoch.
\item For each initiator \(i\) in fixed order, sample a neighbor \(h\) and one eligible friend-of-friend \(k\) from \(\mathcal N_h\setminus(\mathcal N_i\cup\{i\})\); skip if no candidate exists.
\item Compute \(f_i\) before the proposal. If both endpoint degrees are below \(\Dmax\), test addition of \((i,k)\) and accept it only for a strict fitness increase.
\item Otherwise, examine the initiator's existing edges \((i,j)\): require degrees of \(i,j\) above one, candidate \(k\) below \(\Dmax\), and a common neighbor of \(i\) and \(j\). Evaluate feasible one-edge replacements \((i,j)\mapsto(i,k)\).
\item Commit only the highest-scoring strictly improving feasible swap, if one exists; otherwise retain the current edges. Each initiator commits at most one operation before the next begins.
\end{enumerate}
This is a summary of the reference serialized procedure~\citep{liang2026lfhe}; it does not represent an asynchronous or globally optimal rewiring algorithm.

The acceptance test is initiator-local: a direct addition requires both endpoint degrees below \(\Dmax\) and a strict increase in \(f_i\). If addition is unavailable, feasible one-edge swaps are checked against an online candidate endpoint below threshold and a local common-neighbor safeguard for the removed edge; the best strictly improving swap is committed. The common-neighbor test checks local redundancy, not all possible concurrent failures. The threshold \(\Dmax\) regulates online edge acquisition. Because the connected Erd\H{o}s--R\'enyi initializer is specified by expected degree, some nodes may initially exceed the threshold; these degrees are preserved unless subsequently changed by topology updates. Consequently, fixed-\(\Dmax=4\) runs have mean degree near \(4.4\), while endpoint maximum degree can exceed four.

\section{Experimental Details}
\label{app:details}

Controlled scaling uses CIFAR-10~\citep{krizhevsky2009learning},
class-wise Dirichlet \(\alpha=0.3\),
\(N\in\{10,50,100,200\}\), seeds \(\{42,43,44\}\), 300 communication
rounds, one local epoch per round, SGD with learning rate \(0.05\),
batch size \(32\), evaluation every five rounds, and LFHE topology
updates every five rounds.

All methods use the same CIFAR-10 convolutional model: two convolutional stages with 32 and 64 channels, batch normalization and max pooling, followed by a 256-dimensional fully connected hidden representation, dropout \(0.3\), and a 10-way classifier. In the controlled-scaling runs, LFHE uses the 10-dimensional class-wise mean of the final classifier weight matrix as its topology representation signal. The separate matched-FoF diagnostic uses the complete flattened \(10\times256\) final classifier weight matrix. The architecture, preprocessing pipeline, and optimizer configuration are held identical across Ring, Static Random, and LFHE. The controlled conditions vary data allocation and communication topology; because training uses one local epoch, the data regime also determines the per-round local SGD workload.

Ring uses a cycle. Static Random and LFHE use seed-matched connected Erd\H{o}s--R\'enyi initializations with expected degree four. For neighboring clients,
\[
a_{ij}=\frac{1}{1+\max(d_i,d_j)},\qquad
a_{ii}=1-\sum_{j\in\cN_i}a_{ij}.
\]

Under fixed-total training, all 50,000 CIFAR-10 training examples are partitioned across the population. Under fixed-per-client training, each client draws 500 local-dataset indices only from its own Dirichlet pool and reuses the resulting local dataset throughout training. This holds the nominal number of local training entries fixed at 500 per client. Sampling is without replacement when the client's pool contains at least 500 indices and with replacement otherwise; repeated indices therefore do not introduce data from another client's pool. Because the number of \emph{distinct} examples per client is not separately tabulated, the fixed-per-client protocol should not be read as fixing distinct data availability. At the population-average level, fixed-total allocations correspond to about 1,000 versus 250 entries per client at \(N=50\) versus \(N=200\), or approximately 32 versus 8 mini-batches per local epoch (batch size 32); fixed-per-client uses roughly 16 mini-batches at either population size. Heterogeneous client pools yield variation around these illustrative averages.

The two LFHE threshold regimes are
\[
\text{fixed: }\Dmax=4,
\qquad
\text{scaled: }\Dmax=\max(2,\lfloor2\ln N\rfloor)
=\{4,7,9,10\}.
\]

The controlled fixed-threshold runs store exact graph snapshots every five communication rounds together with the final topology. Algebraic connectivity is computed from the unweighted combinatorial Laplacian \(L=D-A\) of these exact edge lists. This diagnostic is \(\lambda_2(L)\), not \(\rho=\lVert P-J\rVert_2\) of the Metropolis matrix; we do not report a Metropolis spectral time series. Historical scaled-threshold runs without evolving adjacency snapshots are not used to construct spectral trajectories.

Model disagreement is
\[
\mathcal D_t
=
\frac{1}{N}\sum_{i=1}^{N}
\left\lVert
\theta_{i,t}-\bar{\theta}_t
\right\rVert_2^2,
\qquad
\bar{\theta}_t
=
\frac{1}{N}\sum_i\theta_{i,t}.
\]
For each setting we report the mean of the recorded test-accuracy endpoints across three paired seeds and their sample SD. All uncertainty values reported for the controlled study are sample SD over the three paired seeds.

\begin{table}[H]
\centering
\small
\caption{Previously published matched-budget FoF diagnostic reproduced from \citet[Appendix H, Table~8]{liang2026lfhe}, not a new control under the scaling protocol. Runs use CIFAR-10, \(N=30\), \(\alpha=0.1\), seeds 42--44, 300 rounds, topology updates every five rounds, \(\Dmax=6\), and six proposals per update. \(\Delta\lambda_2\) uses the shared initial graph; mean divergence averages the final 20\% of evaluation rounds. Mean rewires counts accepted operations per topology update.}
\label{tab:matched}
\begin{tabular}{lccccc}
\toprule
Method &
Final \(\lambda_2\) &
\(\Delta\lambda_2\) &
Final Acc. &
Mean Div. &
Mean Rewires \\
\midrule
Static
& 0.592 & 0.000 & 0.6675 & 0.766 & 0.000 \\
Similarity-only
& 0.578 & -0.014 & 0.6704 & 0.727 & 0.183 \\
Exploration-only
& 1.038 & 0.447 & 0.7356 & 0.221 & 3.406 \\
Structural-only
& 2.306 & 1.715 & 0.7508 & 0.136 & 1.111 \\
LFHE
& 2.247 & 1.655 & 0.7482 & 0.133 & 1.333 \\
\bottomrule
\end{tabular}
\end{table}

As reported by \citet{liang2026lfhe}, Exploration-only accepts more topology operations per update than Structural-only or LFHE but reaches lower accuracy and algebraic connectivity. Structural-only slightly exceeds full LFHE in mean accuracy (0.7508 versus 0.7482) and connectivity (2.306 versus 2.247); the combined score is not uniformly optimal. This externally reported method diagnostic has a different \(N\), heterogeneity, representation, and proposal budget from the present scaling runs. It provides methodological context, not new evidence that topology adaptation alone causes the performance differences in Section~\ref{sec:results}.

\section{Controlled Scaling Results}
\label{app:controlled}

\begin{table}[H]
\centering
\small
\caption{Final fixed-per-client accuracy (\%, mean $\pm$ sample SD over seeds 42--44).}
\label{tab:accuracy}
\begin{tabular}{lcccc}
\toprule
Method & $N=10$ & $N=50$ & $N=100$ & $N=200$ \\
\midrule
Ring
& 61.34$\pm$0.77
& 56.00$\pm$1.43
& 51.77$\pm$0.57
& 44.46$\pm$0.27 \\
Static Random
& 64.95$\pm$1.40
& 66.69$\pm$1.56
& 66.10$\pm$0.43
& 63.24$\pm$0.50 \\
LFHE, fixed ($\Dmax=4$)
& 65.43$\pm$1.92
& 70.52$\pm$0.07
& 68.81$\pm$0.35
& 66.84$\pm$0.27 \\
LFHE, scaled threshold
& 65.43$\pm$1.92
& 72.49$\pm$0.77
& 72.31$\pm$0.55
& 72.04$\pm$0.46 \\
\bottomrule
\end{tabular}
\end{table}

The 0.07 sample SD for fixed-$\Dmax$ LFHE at $N=50$ is computed
directly across the three completed seeds.

\begin{table}[H]
\centering
\small
\caption{Final fixed-$\Dmax=4$ LFHE endpoints at $N\in\{50,100,200\}$ (mean $\pm$ sample SD).}
\label{tab:fixed4}
\begin{tabular}{llccc}
\toprule
Regime & Metric & $N=50$ & $N=100$ & $N=200$ \\
\midrule
\multirow{3}{*}{Fixed-total}
& Accuracy (\%) & 74.65$\pm$1.89 & 68.64$\pm$0.59 & 60.51$\pm$0.96 \\
& Disagreement  & 0.474$\pm$0.116 & 0.354$\pm$0.004 & 0.257$\pm$0.029 \\
& $\lambda_2$   & 0.603$\pm$0.142 & 0.465$\pm$0.124 & 0.523$\pm$0.063 \\
\midrule
\multirow{3}{*}{Fixed-per-client}
& Accuracy (\%) & 70.52$\pm$0.07 & 68.81$\pm$0.35 & 66.84$\pm$0.27 \\
& Disagreement  & 0.348$\pm$0.137 & 0.445$\pm$0.027 & 0.557$\pm$0.030 \\
& $\lambda_2$   & 0.550$\pm$0.217 & 0.426$\pm$0.076 & 0.497$\pm$0.058 \\
\bottomrule
\end{tabular}
\end{table}

\begin{table}[H]
\centering
\small
\caption{Final fixed-per-client model disagreement.}
\label{tab:disagreement}
\begin{tabular}{lcccc}
\toprule
Method & $N=10$ & $N=50$ & $N=100$ & $N=200$ \\
\midrule
Ring
& 0.439 & 27.482 & 54.391 & 67.414 \\
Static Random
& 0.129 & 0.578 & 0.611 & 0.803 \\
LFHE, fixed ($\Dmax=4$)
& 0.180 & 0.348 & 0.445 & 0.557 \\
LFHE, scaled threshold
& 0.180 & 0.110 & 0.095 & 0.123 \\
\bottomrule
\end{tabular}
\end{table}

\section{Communication Accounting}
\label{app:communication}

\begin{table}[H]
\centering
\small
\caption{LFHE cumulative directed model transmissions under fixed-per-client training (mean over seeds 42--44).}
\label{tab:lfhecomm}
\begin{tabular}{lccc}
\toprule
Threshold regime & $N=50$ & $N=100$ & $N=200$ \\
\midrule
Fixed ($\Dmax=4$)
& 66,710 & 131,419 & 265,519 \\
Scaled threshold
& 98,160 & 247,254 & 541,831 \\
\midrule
Increase
& 47.1\% & 88.1\% & 104.1\% \\
\bottomrule
\end{tabular}
\end{table}

\begin{table}[H]
\centering
\small
\caption{Fixed-per-client communication endpoints at $N=100$ and $N=200$. Counts include directed model transmissions and exclude separately unlogged topology-control traffic.}
\label{tab:n100comm}
\begin{tabular}{lcc}
\toprule
Method & $N=100$ & $N=200$ \\
\midrule
Ring                         & 60,000  & 120,000 \\
Static Random                & 120,600 & 242,000 \\
LFHE, fixed ($\Dmax=4$)      & 131,419 & 265,519 \\
LFHE, scaled threshold       & 247,254 & 541,831 \\
\bottomrule
\end{tabular}
\end{table}

At $N=100$ and $N=200$, fixed-threshold LFHE uses approximately 9\% and 9.7\% more model transmissions than Static Random, respectively. Scaling the LFHE threshold from four to nine at $N=100$ increases communication by 88.1\%, while scaling it from four to ten at $N=200$ increases communication by 104.1\%.

\section{Graph--Model Dynamics Under a Fixed Degree Budget}
\label{app:dynamics}

\begin{figure}[H]
\centering
\includegraphics[width=\linewidth]{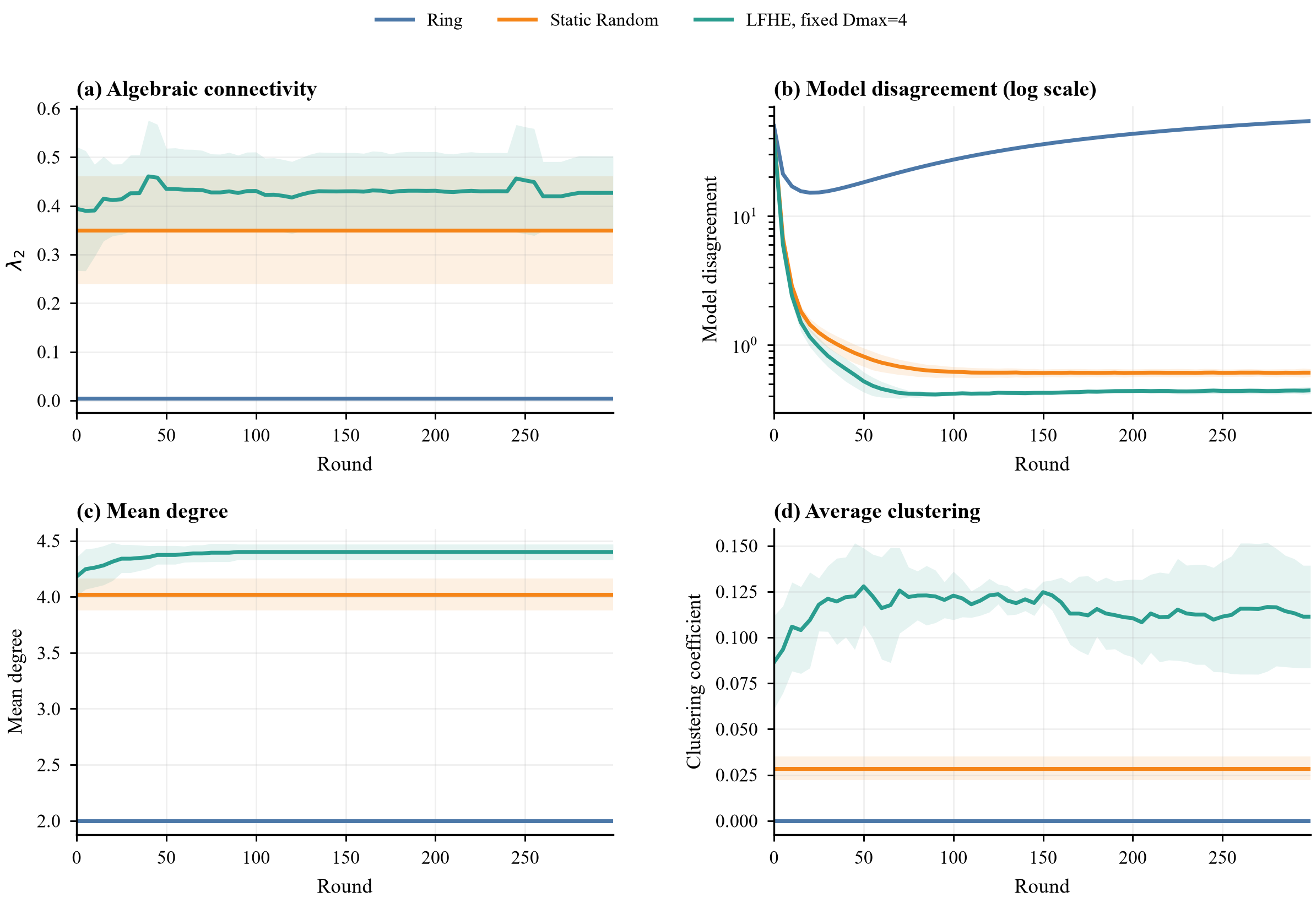}
\caption{\small Graph--model dynamics at \(N=100\) under fixed-per-client training with \(\Dmax=4\). LFHE maintains stronger algebraic connectivity than Ring while mean degree remains near \(4.4\), and its disagreement converges to the low-disagreement regime shared with Static Random. Curves show means over seeds 42--44; shading is sample SD.}
\label{fig:dynamics}
\end{figure}

At \(N=100\), LFHE's algebraic connectivity changes from \(0.394\) to \(0.426\) while its mean degree remains near \(4.4\). Static Random remains near \(0.35\), whereas the Ring has substantially weaker algebraic connectivity. At the same time, LFHE's model disagreement contracts from a large initial value to \(0.445\), Static Random stabilizes near \(0.6\), and Ring enters a substantially higher-disagreement regime.

Structural and optimization changes are concentrated in the early part of training, after which both topology and disagreement become comparatively stable. The exact trajectories show that topology evolution and consensus formation occur on related but non-identical time scales, reinforcing the view of LFHE as a coupled graph--model process rather than a single spectral optimization step.

\section{Degree-Budget Diagnostic}
\label{app:degree}

\begin{figure}[H]
\centering
\includegraphics[width=\linewidth]{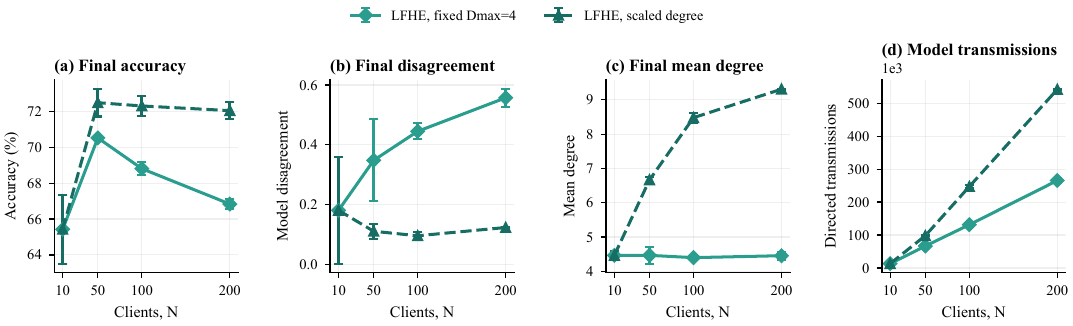}
\caption{\small LFHE degree-budget sensitivity under fixed-per-client training. Increasing the online threshold produces higher endpoint accuracy and lower disagreement, together with higher realized degree and communication.}
\label{fig:degreeappendix}
\end{figure}

At \(N=50\), changing \(\Dmax\) from 4 to 7 raises accuracy from \(70.52\%\) to \(72.49\%\), reduces disagreement from \(0.348\) to \(0.110\), and increases transmissions from \(66{,}710\) to \(98{,}160\). At \(N=100\), changing \(\Dmax\) from 4 to 9 raises accuracy from \(68.81\%\) to \(72.31\%\), reduces disagreement from \(0.445\) to \(0.095\), and increases transmissions from \(131{,}419\) to \(247{,}254\). At \(N=200\), changing \(\Dmax\) from 4 to 10 raises accuracy from \(66.84\%\) to \(72.04\%\), reduces disagreement from \(0.557\) to \(0.123\), and increases transmissions from \(265{,}519\) to
\(541{,}831\).

This comparison shows LFHE's degree-budget sensitivity; communication-normalized results appear in Appendix~\ref{app:commnorm}.

\section{Communication-Normalized Learning}
\label{app:commnorm}

\begin{figure}[H]
\centering
\includegraphics[width=0.96\linewidth]{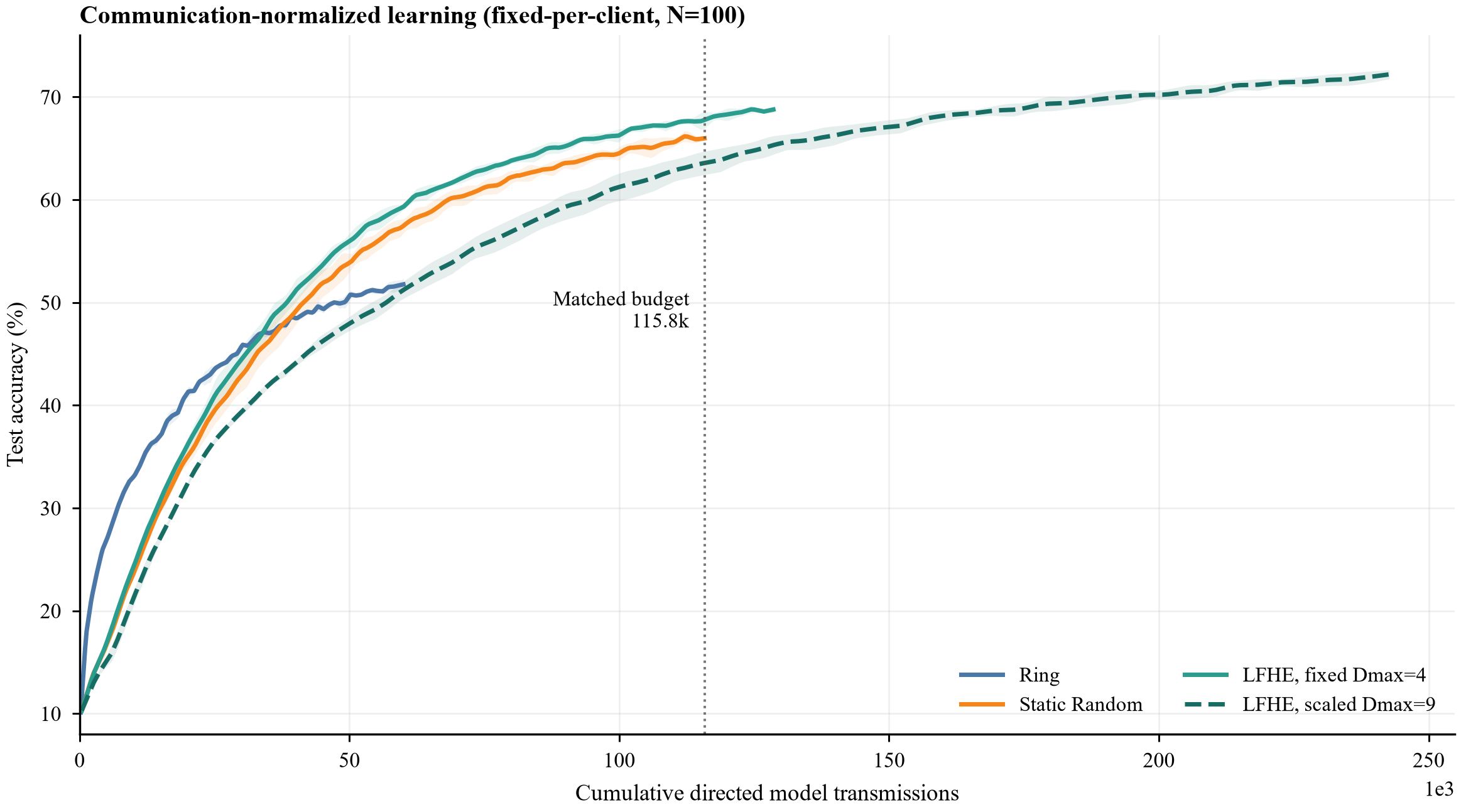}
\caption{\small Accuracy versus cumulative directed model transmissions at \(N=100\) under fixed-per-client training. Scaled-threshold LFHE attains the highest endpoint after extending substantially beyond the fixed-threshold communication range.}
\label{fig:commappendix}
\end{figure}

At a common budget of \(115.8\)k transmissions, Static Random, fixed-\(\Dmax\) LFHE, and scaled-threshold LFHE achieve \(65.98\%\), \(67.77\%\), and \(63.58\%\), respectively. The transmissions required to first reach \(65\%\) accuracy are approximately \(106.3\)k, \(88.6\)k, and \(130.7\)k. Fixed-threshold LFHE therefore reaches stronger learning within the shared communication range, whereas scaled-threshold LFHE trades a larger total budget for a higher eventual endpoint.

\section{Scope and Extensions}
\label{app:limits}

The controlled study focuses on CIFAR-10, populations up to \(N=200\), and three paired seeds, with communication measured through directed model transmissions. The fixed-total and fixed-per-client regimes do not hold local SGD steps or numbers of distinct samples constant; this prevents attributing the observed protocol effect solely to data dilution. Within this setting, the experiments characterize how data allocation, mixing structure, local search visibility, and communication capacity shape the observed scaling regime. LFHE uses final classifier weights as its representation signal throughout; evaluating alternative representation layers or learned signals would further characterize the sensitivity of the topology process.

Extending the analysis to larger populations and broader learning tasks would test whether the same graph--model regimes persist beyond the finite-population setting studied here. An equal-SGD-step comparison with distinct-example counts would isolate the data-allocation mechanism more cleanly. A static graph matched to LFHE's realized degree and transmission schedule would better separate connectivity capacity from adaptive selection, and computing \(\rho_t\) from stored graph snapshots would align the empirical mixing diagnostic with the contraction inequality. Concurrent topology updates would extend the present sequential protocol toward asynchronous decentralized execution. Comparisons with additional adaptive-topology processes such as PENS and Morph are also natural extensions under the same controlled data and communication regimes.
\end{document}